\documentclass{article}
\usepackage[pdftex,dvipsnames]{xcolor}
\usepackage[utf8]{inputenc}
\usepackage{authblk}
\usepackage{amssymb}
\usepackage{amsmath}
\usepackage{mathtools}
\usepackage{amsfonts}
\usepackage{caption}
\usepackage{subcaption}
\usepackage{graphicx}
\usepackage[T1]{fontenc}
\usepackage{soulutf8}
\usepackage{makeidx}
\usepackage{url}
\usepackage{pbox}
\usepackage{diagbox}
\usepackage{float}
\usepackage{multirow, makecell}
\usepackage[a4paper]{geometry}
\usepackage{times}
\usepackage{todonotes}
\usepackage{ctable}
\usepackage{soul}
\usepackage{standalone}
\graphicspath{{img/}}

\newcommand{\vect}[1]{\boldsymbol{#1}}

\begin{document}

    \title{Classification and Self-Supervised Regression of Arrhythmic ECG Signals Using Convolutional Neural Networks}
    \author[a]{Bartosz Grabowski}
    \author[a,*]{Przemysław Głomb}
    \author[b]{Wojciech Masarczyk}
    \author[c,a]{Paweł Pławiak}
    \author[d]{Özal Yıldırım}
    \author[e,f,g,h,i]{U Rajendra Acharya}
    \author[j,k]{Ru-San Tan}

    \affil[a]{Institute of Theoretical and Applied Informatics, Polish Academy of Sciences, Bałtycka 5, 44-100 Gliwice, Poland}
    \affil[b]{Warsaw University of Technology, Pl. Politechniki 1, 00-661 Warsaw, Poland}
    \affil[c]{Department of Computer Science, Faculty of Computer Science and Telecommunications, Cracow University of Technology, Warszawska 24, 31-155 Krakow, Poland}
    \affil[d]{Department of Software Engineering, Firat University, Elazig, Turkey}
    \affil[e]{School of Science and Technology, Singapore University of Social Sciences, Singapore}
    \affil[f]{School of Business (Information Systems), Faculty of Business, Education, Law and Arts, University of Southern Queensland, Australia}
    \affil[g]{School of Engineering, Ngee Ann Polytechnic, 535 Clementi Road, 599489, Singapore}
    \affil[h]{Department of Bioinformatics and Medical Engineering, Asia University, Taiwan}
    \affil[i]{Research Organization for Advanced Science and Technology (IROAST), Kumamoto University, Kumamoto, Japan}
    \affil[j]{Department of Cardiology, National Heart Centre Singapore, 169609, Singapore}
    \affil[k]{Duke-NUS Medical School, 169857, Singapore}
    \affil[*]{Corresponding author: Przemysław Głomb, przemg@iitis.pl}

    \date{}
    \maketitle

    \begin{abstract}{}
    Interpretation of electrocardiography (ECG) signals is required for diagnosing cardiac arrhythmia. Recently, machine learning techniques have been applied for automated computer-aided diagnosis. Machine learning tasks can be divided into regression and classification. Regression can be used for noise and artifacts removal as well as resolve issues of missing data from low sampling frequency. Classification task concerns the prediction of output diagnostic classes according to expert-labeled input classes. In this work, we propose a deep neural network model capable of solving regression and classification tasks. Moreover, we combined the two approaches, using unlabeled and labeled data, to train the model. We tested the model on the MIT-BIH Arrhythmia database. Our method showed high effectiveness in detecting cardiac arrhythmia based on modified Lead II ECG records, as well as achieved high quality of ECG signal approximation. For the former, our method attained overall accuracy of $87.33\%$ and balanced accuracy of $80.54\%$, on par with reference approaches. For the latter, application of self-supervised learning allowed for training without the need for expert labels. The regression model yielded satisfactory performance with fairly accurate prediction of QRS complexes. Transferring knowledge from regression to the classification task, our method attained higher overall accuracy of $87.78\%$.
    \end{abstract}

    \section{Introduction}

Electrocardiography (ECG) reads out a spatial map of the time-varying electrical potentials of the heart acquired using electrodes placed at specific locations on the surface of the body. Interpretation of the ECG unveils structural and functional abnormalities of the heart that can aid the noninvasive diagnosis of cardiovascular diseases~\cite{KUNDU_2000, LUZ_2016, Zuo_2008}. Importantly, ECG is the most important diagnostic tool for arrhythmia detection~\cite{Owis_2002, Ozal2018, Plawiak_2020, Hannun_2019, Thakor_1991}. As the abnormal heart beats often occur sporadically and are not present at all times, ECG recordings may have to be carried out repeatedly or continuously over an extended period of time, e.g., days with ambulatory Holter devices~\cite{LUZ_2016}. Due to the high signal data volume, manual interpretation is time-consuming and susceptible to fatigue-induced error. This has spurred the introduction of automated computer-aided diagnostic systems, which may be based on machine learning. Some machine learning techniques are able to evaluate individual heartbeat signals on ECG records~\cite{Hong_2020} to complete tasks like classification, localization and prediction.
Among the many explored applications of machine learning for ECG signal analysis, two general problems stand out: regression and classification. Regression is a quantitative prediction task that maps the input data into output consisting of real or continuous values. For ECG data, regression problem can take various forms, including segmentation method for detecting ECG P, Q, R, S, and T waves~\cite{Aspuru_2019}; reference method for removing noise artifacts from ECG signals~\cite{Dora_2016}; and increasing the spatial resolution of ECG through lead prediction~\cite{Nallikuzhy_2015}. On the other hand, classification is a predictive technique that maps the input data to output data (targets, classes or categories) to arrive at the correct class labels to which the input should belong. Examples of works published on ECG dataset classification include labeling heartbeats into one of the five beat classes according to the ANSI/AAMI EC57:1998 standard~\cite{Chazal_2004}; classification ECG segments into normal and multiple arrhythmia classes~\cite{Acharya2017}; and classification of myocardial infarction~\cite{Baloglu2019}. In many situations encountered in automated analyses, regression and classification tasks are intertwined, as the former can be used to enhance the performance of the latter, e.g., by mitigating ECG degradation from noise and artifacts as well as missing data from low sampling frequency~\cite{Clifford_2006}. 
In this article, we present a novel approach of ECG signal modeling that uses a convolutional neural network (CNN) for both regression and classification. One of the advantages of our method is the flexibility of neural networks, making it possible to adapt a single neural network architecture for multiple tasks. We have exploited this trait to develop a single neural network model that is capable of modeling large parts of the input ECG signal as well as classifying the same data. Moreover, unlike classification, which requires training the model on expert-annotated ECG signal data, the model can accomplish the regression task without the need for data labeling. What is also worth noting, the knowledge gained from the self-supervised regression task can be seamlessly transferred to the downstream classification task. This two-pronged approach offers optionality that may improve diagnostic classification at little additional computational cost. On the other hand, the disadvantage of our approach is the relatively complex method with a lot of hyperparameters that needs tuning, which can require a lot of time and computational resources to optimize. In summary, the novelty of our work is as follows:
\begin{enumerate}
    \item We propose an algorithm that is able to achieve good results at two different tasks, regression and classification, both of which are important components for development of automated computer-aided systems for ECG signal analysis.
    \item Combining within the same CNN model the dual tasks of predicting parts of the ECG signal (regression) and detection of cardiac arrhythmia (classification) ―which use unlabeled and labeled data, respectively, for training― may improve model performance without increasing model complexity inordinately.
\end{enumerate}
The paper is structured as follows. Section~\ref{sec:related-work} presents the related work. Section~\ref{section:method} introduces materials and methods, and Section~\ref{section:experiments} summarizes and discusses the results of the conducted experiments. Section~\ref{sec:conclusions} concludes the paper.

    \subsection{Related work}\label{sec:related-work}
    
        \paragraph{Classification of ECG signals}

The process of classification of ECG signals is traditionally split into feature extraction and classification steps. The feature vectors obtained in the feature extraction stage are fed into classifiers. In automated ECG classification, features such as morphology and intervals of specific waves on the ECG signal have been widely used in the past. In~\cite{Yeap_1990}, QRS wave width, amplitude, and offset, T-wave slope, and prematurity features were obtained for each beat, and the classification was made using a neural network. In~\cite{Chazal_2004}, the authors proposed a detection approach using morphologic features, such as QRS wave onset/offset and T-wave offset, heartbeat intervals, e.g., RR intervals, features for automated ECG beat classification using linear discriminant-based classifier models. Features may also be transformed using methods like Fourier and wavelet transformation prior to classification. In~\cite{Dokur_2001}, the authors obtained distinctive features for ECG signals using the Fourier and wavelet transform approach. With a hybrid neural network model, they reported 96\% accuracy in ten-class ECG beat recognition using these transformed features. In another study~\cite{YILDIRIM_2019} that used discrete wavelet transform, the features obtained from the coefficients of different wavelet levels were classified using extreme learning machines architecture. In~\cite{ENGIN_2004}, the authors used auto-regressive model coefficients, third-order cumulants, and discrete wavelet transform approaches to extract ECG signal feature vectors, which were classified using fuzzy-hybrid neural networks. Various statistical methods can be used for feature extraction, including principal component analysis, linear discriminant analysis, independent component analysis, higher order statistic, and transformation methods. Martis et al.~\cite{MARTIS_2013} used principal component, linear discriminant, and independent component analyses in the feature extraction step of their ECG beat classification model. The authors in~\cite{Osowski_2001} performed ECG beat classification by feeding second, third and fourth order features to a hybrid fuzzy neural network with higher order statistic. Acharya et al.~\cite{ACHARYA_2017} extracted higher order statistic bispectrum and cumulant features from ECG signals to classify coronary artery disease using a k-nearest neighbor (KNN) classifier. As some feature extraction methods can generate large-sized feature vectors, various approaches can be used to select the most distinctive features to reduce the dimensionality. In~\cite{Plawiak_2018}, extracted features were selected using genetic algorithm, which were then fed to a support vector machine (SVM)-based classifier. In~\cite{Mar_2011}, a sequential forward floating search was used to select features, which were then classified using the multilayer perceptron approach.

Several authors have used deep models for analysis of physiological signals, including for ECG arrhythmia classification~\cite{Faust2018, Murat2020, Gupta_2021}, and the number of publications on deep learning models has increased significantly in the last few years~~\cite{Faust2018}. In~\cite{Acharya2017}, the authors developed a CNN to classify normal and arrhythmia ECG segments that did not require QRS detection. The 12-layer CNN model was also used by~\cite{Wu_2021} to classify the five micro-classes of heartbeat types. In~\cite{Ozal2018}, the authors proposed a one-dimensional (1D)-CNN to process long-duration ECG signal fragments, which was computationally efficient and highly accurate. The authors of~\cite{Baloglu2019} used a CNN to classify myocardial infarction using 12-lead ECG signals, and achieved 99\% accuracy. Long short-term memory (LSTM) network was used by~\cite{Faust2018-2} to detect atrial fibrillation based on heart rate signals. The authors partitioned the data with a sliding window of 100 beats and used the resulting signal segments to train and evaluate the network. In~\cite{Yildirim2019}, the authors used a convolutional auto-encoder to reduce the ECG signal size and a LSTM network to process the compressed data for arrhythmia detection. Compression of the signal minimized storage requirement and data transfer costs, and was able to reduce training time of the LSTM network without significantly compromising model performance. In~\cite{Lih2020}, the authors combined CNN and LSTM network for three-class classification of coronary artery disease, myocardial infarction and congestive heart failure using ECG data. Authors of~\cite{Smigiel_2021} proposed three different neural network architectures for classification of ECG signals and obtained the best results for the architecture containing entropy features, while the one without it had the highest computational efficiency.


        \paragraph{Regression of ECG signals}

Regression techniques can be used to rectify ECG signal issues, such as noise and artifacts, as well as missing data from low sampling frequency~\cite{Clifford_2006}, which may affect performance in arrhythmia classification. Various techniques can be used for  removing noise and artifacts in ECG signals, including regression, interpolation, and deep learning. In~\cite{Rheinberger_2008}, the authors used a regression-based model to remove motion artifacts caused by cardiopulmonary resuscitation in the ECG signal output of automatic external defibrillators to improve the rhythm detection algorithm. Sidek et al~\cite{SIDEK_2013} used cubic Hermite and piecewise cubic spline interpolation approaches to improve ECG signal quality, and reported that the improved quality of the signals conferred higher performance for biometric matching. Similarly, Kamata et al.~\cite{Kamata_2016} proposed a just-in-time interpolation approach to reduce signal artifacts to facilitate accurate R wave detection, which is of fundamental importance for heart rate variability analysis. Apart from denoising for signal enhancement, Nallikuzhy et al.~\cite{Nallikuzhy_2015} proposed a multiscale linear regression model that was able to increase ECG spatial resolution. The regression approach is also used in segmentation tasks. Aspuru et al.~\cite{Aspuru_2019} used linear regression to parse the P, Q, R, S and T wave regions of ECG signals for downstream classification.

Autoencoder algorithms, a deep learning method, have also been used in the compression and reconstruction as well as denoising of ECG signals. Yildirim et al.~\cite{YILDIRIM_2018csr} proposed a deep autoencoder  that could reduce the original ECG signal input size to improve model efficiency, and reconstruct the original ECG signal at the output. These structures have also been used to denoise ECG signals and enhance their quality~\cite{XIONG_2016, Chiang_2019}. In~\cite{XIONG_2016}, the authors designed a denoising autoencoder model with wavelet transform for noise removal. Recurrent structures have also been actively used for ECG denoising~\cite{Rahman_2012}. Generative adversarial networks (GAN) have also been frequently employed for similar purposes. Antczak~\cite{Antczak_2020} used the GAN approach to generate synthetic ECG signals, which were used for training noise removing models. Golany et al.~\cite{Golany_2020} used the synthetic data produced by the GAN approach to increase the accuracy of heartbeat classification.

        
        \paragraph{Self-supervised learning}

Self-supervised learning strategies have recently gained popularity in machine learning-based diagnosis of medical images~\cite{Chen_2019,Jamaludin_2017}, electroencephalography signals (EEG)~\cite{Banville_2020,Zhao_2020,Banville_2020a}, and ECG signals~\cite{Sarkar_2020, Mehari_2022}. This learning strategy can learn from unlabeled data and does not require a supervisor annotated dataset. While self-supervised learning strategy per se may not improve the accuracy significantly compared with labeled data learning, it has some important advantages~\cite{hendrycks2019using}. Self-supervised methods have a structure that can self-learn through data without the need for class labels. As such, they circumvent the need to annotate large amounts of data by experts in conventional deep learning. This is especially useful because of the limitations of computational resources and the scarcity of available data for research~\cite{Banville_2020, Chen_2019}.

Many prior works on self-supervised learning have been in the field of natural language processing. In~\cite{Dai2015}, two approaches to utilize unlabeled data for pretraining of the recurrent neural networks were tested. The first approach was to predict what comes next in a sequence, while the second used an autoencoder to learn effective encoding of the input sequence. The authors demonstrated that both approaches helped stabilize the training as well as improved generalization of the LSTM. In~\cite{Howard2018}, the authors proposed a universal language model fine-tuning method that utilized both general-domain as well as target task-specific language models to pretrain the LSTM model, which was then fine-tuned for the target text classification task. They proposed novel fine-tuning techniques to prevent catastrophic forgetting as well as enable robust learning. In~\cite{Radford2018}, the authors used a transformer architecture~\cite{Vaswani2017} to solve multiple natural language understanding tasks. The model was first pretrained on a language modeling task using a large unlabeled dataset. Next, the network was fine-tuned to solve one of many specific target tasks, using task-specific input transformations where necessary. In~\cite{Radford2019, Brown2020}, transformer language models were shown to be able to learn the target tasks without being explicitly trained to perform them. The networks were thus able to work in a fully unsupervised fashion: learning to perform other natural language tasks after being trained to do a language modeling task. In~\cite{Devlin2018}, the authors proposed a deep bidirectional transformer model, which was pretrained on two different unsupervised tasks: masked language modeling, where some percentage of the input was hidden at random and the task was to predict the hidden parts; and the next sentence prediction task, where the model must predict if the second received sentence was the one following the first sentence. In~\cite{Chen2020}, the authors applied the unsupervised pretraining methodology that was common in natural language processing to computer vision. They performed ImageNet classification in three steps: unsupervised pretraining, which was based on the approach presented in~\cite{Chen2020-2}; supervised fine-tuning; and distillation with unlabeled data.

In this work, we use the same dataset preparation steps as in~\cite{Ozal2018}. Moreover, we utilize self-supervised learning, which was used in e.g.~\cite{Sarkar_2020, Mehari_2022} to improve the classification accuracy of our model.

    \section{Material and methods}\label{section:method}
    
    We designed a CNN that could accomplish the dual tasks of ECG beat regression and ECG
segment arrhythmia classification. First, the network architectural requirements were posed as
an optimization problem, which we solved by using Ray Tune library~\cite{Liaw2018} to
choose the best performing configurations. Neural networks with different permutations of
user-defined hyperparameters were trained on the regression followed by classification
tasks. The value of validation loss of individual architectures with specific combinations of hyperparameter settings was expressed as scores. The aim of the optimization process
was to find models with the lowest score for the classification task. For optimization training, the
following options for hyperparameter settings were considered: batch size, 50 and 100;
the number of convolutional layers, 3, 5 and 7 (these were common to both regression and
classification tasks); the number of channels in the first layer, 8 and 16 (this number was
doubled in every successive layer until the maximum, 128); kernel size of the first layer, 64
and 128 (this number was halved in every successive layer until the minimum, 2); kernel
size of max-pooling layers, 3 and 4; inclusion of batch normalization layer, yes and no;
number of classification layers, 1 and 3 (which were added in the classification task);
the number of neurons in the classification layer(s) or the last layer in the case of regression,
1000 and 3000; inclusion of residual connections from all convolutional layers to the first
classification layer, yes and no. Async successive halving algorithm scheduler~\cite{Li2018} was used for hyperparameter optimization, which works by evaluating multiple model configurations and dropping not promising ones based on initial training performance. This allows for time and resource effective search given large hyperparameter space. The hyperparameter search was carried out for approximately
12 days. Upon completion of the optimization, the best architectures and training processes
from those evaluated were chosen.
    
        \subsection{Architecture}

\begin{figure}
            \centering
            \includegraphics[width=\linewidth]{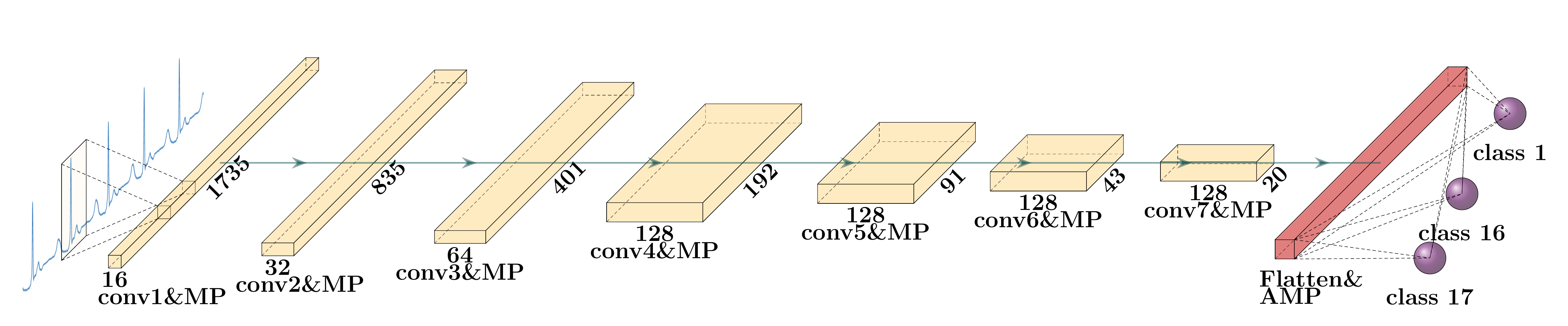}
            \caption{Schema of the optimized network architecture. Each convolution layer (depicted as conv) was followed
by rectified linear unit activation function, batch normalization, and max-pooling layers (depicted as MP). The
numbers depict the number of channels (horizontal) and data dimension (vertical) at
successive layers. After last convolution layer, the data were flattened and adaptive max
pooling (depcited as AMP) was used to obtain the desired size of the data. Depending on the task, differently-shaped
linear layers were added at the end of the network. The figure shows the classification task with output vector of dimension 17. Conv, convolutional layer.}
            \label{fig:network}
        \end{figure}

The optimized CNN architecture (Figure~\ref{fig:network}) comprised seven convolutional layers that were
each followed by rectified linear units activation, batch normalization and max
pooling (pooling size = 4, stride = 2). Starting with a kernel size of 128 in the first
convolutional layer, the values for successive layers were progressively halved until 2 in
the seventh layer. The number of channels in the first layer was 16, which was doubled
with each succeeding layer until a maximum of 128 channels in the fourth through seventh
layers. All convolutional layers operated with stride set to 1. After the penultimate layer, the model employed adaptive max-pooling layer which automatically chooses stride length or padding value for the max-pool operation in order to meet desired output shape for that layer. The difference between
the regression and classification tasks architectures lay in the last layer, a
linear layer that transformed representations from the penultimate layer to the target part
of the ECG signal or arrhythmia class probabilities, respectively, with corresponding output
vectors of dimensions 100 and 17.

        \subsection{Optimization}

To optimize the network for the regression task, we used SmoothL1 loss, which is defined as:
        $$L(x,y)=\frac{1}{n}\sum_{i}z_i,$$
        where $z_i$ is given by:
        $$z_i=\left\{\begin{array}{ll}
        0.5(x_{i}-y_{i})^2, & \text { if }|x_{i}-y_{i}|<1 \\
        |x_{i}-y_{i}|-0.5, & \text { otherwise }
        \end{array}\right.$$
For the optimization algorithm, we chose Adam with a learning rate = $0.001$ and
$\beta_1 = 0.9$, $\beta_2 = 0.999$. We trained the regression network for 100 epochs with batch size set to
50, and used a standard Kaiming procedure to initialize the weights for both convolutional
and linear layers.
The classification task shared common architecture and optimization parameters with the
regression task except for the different number of final layers' neurons as well as the number of
training epochs, which was 1000. The loss function used in this case was cross-entropy loss.

    \subsubsection{Knowledge transfer}\label{sec:kt}
    
\begin{figure}[!ht]
            \includegraphics[width=\columnwidth]{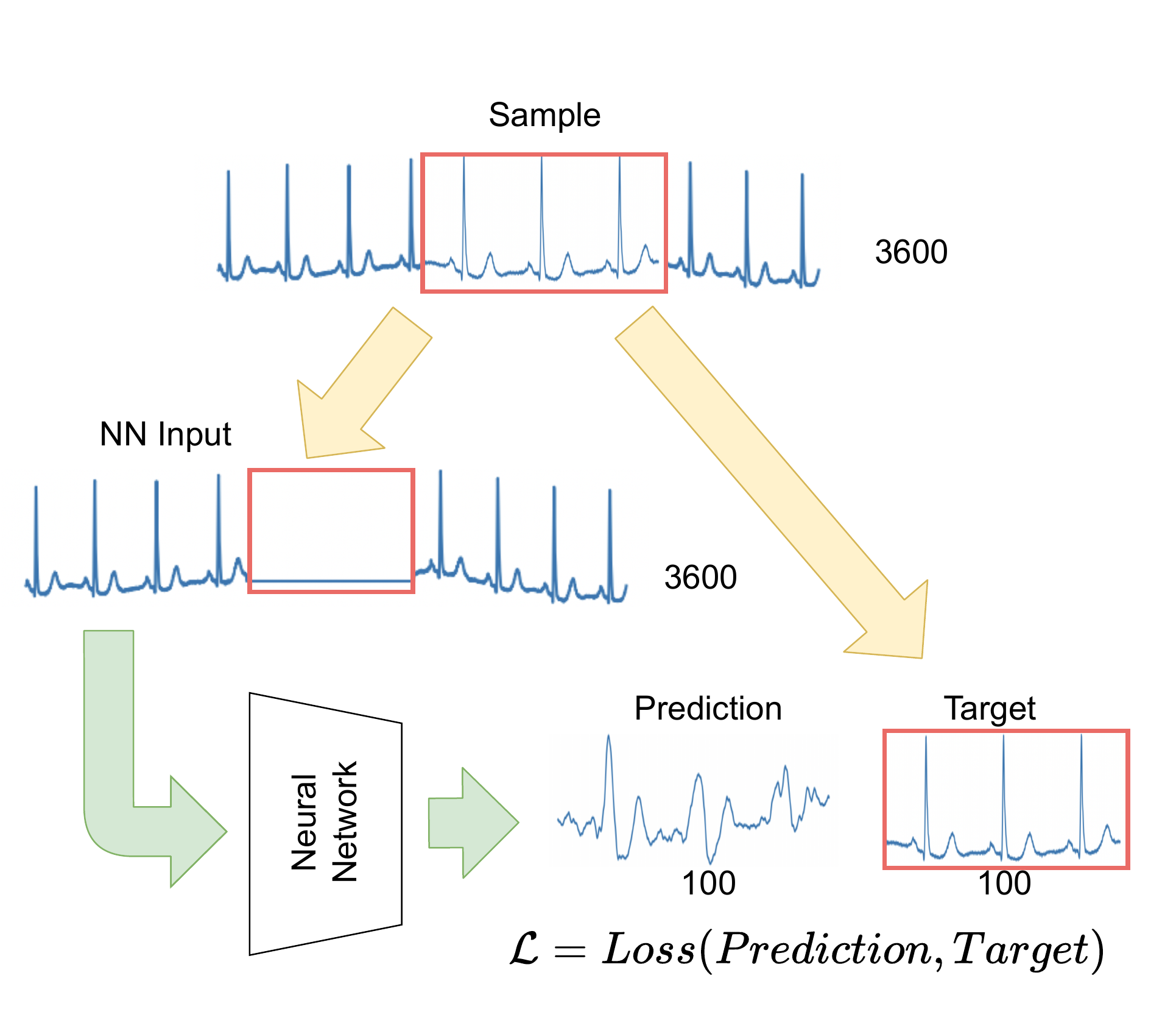}
            \caption{Schema of the prediction task. From a raw 10-second ECG sample of length 3600 randomly drawn from the dataset, a random portion of length 100 comprising ECG beat/s was extracted and its component 100 values zeroed before the whole sample was fed to the neural network (green). The extracted portion constituted the target for the neural network (red). The prediction of the neural network was compared with reference to the target signal (both of length 100), and errors between neural network prediction and target were used to train the neural network. NN -- neural network.}
            \label{figure:prediction}
        \end{figure}

    ECG signals are ubiquitous but the process of annotating the data for supervised learning
requires domain knowledge, and is manually intensive and expensive. Self-supervised
learning can circumvent this limitation, allowing the exploitation of the rich
potential of unlabeled ECG datasets. To this end, we used CNN regression to predict ECG
beats with a dimension size of 100 within individual sample ECG segments. Figure~\ref{figure:prediction} depicts
the process of using self-supervised CNN regression to predict ECG beats within sample
raw ECG segments. The optimized regression model was selected based on the validation loss,
and the results were transferred to the classification task. We postulated that this approach of
transferring knowledge from the first task to the second would improve the performance of
the latter. Details of the experiment in which the output of the regression task was
input to the classification task are discussed in Section~\ref{sec:rtc} below.

    \subsection{Experimental verification}

    \subsubsection{Dataset}\label{section:datasets}

Modified Lead II ECG signals were obtained from the MIT-BIH Arrhythmia~\cite{Moody_2001} database in the PhysioNet~\cite{Goldberger_2000} service. The ECG signals
had been acquired from 19 female (age range: 23 to 89 years) and 26 male subjects (age
range: 32 to 89 years) and comprised 17 classes, including normal sinus rhythm, paced
rhythm and 15 types of other arrhythmias. The ECG signals were recorded at a sampling
frequency of 360 Hz with a gain of 200 analog-to-digital units/mV. For the regression task,
we constructed a dataset DS0 that comprised 99,990 10-second ECG segments to train and
test the CNN. For the classification task, we randomly selected 1,000 non-overlapping 10-
second ECG segments and divided them into two datasets: DS1 for training and validation;
and DS2 for testing~\cite{Plawiak_2018}. Table~\ref{tab:dataset} lists the distribution of the 1,000 ECG segments
from 45 unique subjects by diagnosis (17 classes), and their use for training, validation and
testing. This is the same scheme as the one used in~\cite{Ozal2018}, which allows us to better compare our approach to existing methods. In particular, we use 10-second ECG signal fragments, which is a typical duration of the rhythm strip aquisition on a 12 lead ECG, allowing the usage of our model without complex data processing, e.g. QRS detection. Note that for the rarer diagnoses, the cells contained only one subject or one ECG
segment.
        
        \ctable[
        cap     = Dataset,
        caption = {A description of the database with the selected 1000 ECG segments along with the allocation of signals to the training, validation and test sets.},
        label   = tab:dataset,
        pos     = ht,]
        {llccccc}{}{\FL
            &       & \multicolumn{2}{c}{Number} &\multicolumn{3}{c}{Set}\NN
        No. & Class & Segments & Patients        & Training & Validation & Test\ML
        
        1 & Normal sinus rhythm & 283 & 23 & 200 & 47 & 36 \NN
        2 & Atrial premature beat & 66 & 9 & 44 & 10 & 12 \NN
        3 & Atrial flutter & 20 & 3 & 13 & 3 & 4 \NN
        4 & Atrial fibrillation & 135 & 6 & 96 & 21 & 18 \NN
        5 & Supraventricular tachyarrhythmia & 13 & 4 & 9 & 2 & 2 \NN
        6 & Pre-excitation (WPW) & 21 & 1 & 15 & 4 & 2 \NN
        7 & Premature ventricular contraction & 133 & 14 & 98 & 19 & 16 \NN
        8 & Ventricular bigeminy & 55 & 7 & 38 & 8 & 9 \NN
        9 & Ventricular trigeminy & 13 & 4 & 10 & 2 & 1 \NN
        10 & Ventricular tachycardia & 10 & 3 & 7 & 1 & 2 \NN
        11 & Idioventricular rhythm & 10 & 1 & 7 & 2 & 1 \NN
        12 & Ventricular flutter & 10 & 1 & 6 & 1 & 3 \NN
        13 & Fusion of ventricular and normal beat & 11 & 3 & 7 & 3 & 1 \NN
        14 & Left bundle branch block beat & 103 & 3 & 73 & 11 & 19 \NN
        15 & Right bundle branch block beat & 62 & 3 & 45 & 8 & 9 \NN
        16 & Second-degree heart block & 10 & 1 & 6 & 3 & 1 \NN
        17 & Pacemaker rhythm & 45 & 2 & 26 & 4 & 14 \NN
           & Sum & 1000 & 45 & 700 & 150 & 150 \NN
        }
        
        We created the dataset DS0 to train and test our method on regression scenario. It consists of 99,990 ECG fragments of equal length, which is almost 100 times more than in the case of classification dataset. Each ECG fragment was randomly chosen from among the 45 persons, with each patient having the same number of segments. These patients are the same as in the case of classification dataset. Just as in the case of the classification dataset, each fragment is derived from MLII lead and is 10 seconds long, which means that it contains 3600 samples.

        For each dataset, every ECG fragment was scaled to the $\left[-1,1\right]$ range.

    \subsubsection{Reference methods}\label{sec:ref}

To compare the classification performance of our model, we used two reference methods.
The first, a neural network was a 16-layer 1D-CNN model, which
we had previously employed on 10-second ECG segment data~\cite{Ozal2018}. The
model consisted of seven 1D convolutional layers, four max-pooling layers and two batch normalization
layers, which standardized the inputs to one layer for each mini-batch. The raw ECG segments were first pre-processed using minmax scaler and standard scaler techniques.
For model training, the Adam optimizer with learning rate = 0.001 and decay = 1e-2
parameters was used. ECG arrhythmia classification at the end of the model was performed
using the softmax function. The kernel size, filter size and stride parameters of the 1D
convolutional layers were effective in extracting hierarchical features from long-term ECG
signals. For example, the 1D convolutional layer placed on the first input layer operated
with 128 filters, 50 kernel size and 3 strides on the original input ECG signal~\cite{Ozal2018}.

The second reference method was a standard shallow SVM with
Gaussian kernel. SVM is a supervised method that finds a hyperplane in the feature space
separating the datapoints of different classes by the biggest possible margin. For data that
may not be linearly separable in the original space, the datapoints are first transformed and
kernels used by the SVM classifier to describe the similarity in the morphed space, where it
is supposed that data would be more separate and the hyperplane easier to identify. For
the pre-processing steps, the normalization parameters (e.g., mean estimation) and feature
extraction method were selected through the validation set and not assumed a-priori.
Standard scaling, principal component analysis with $n_{\mathrm{pca}}\in\{3, 10, 50\}$ and fast Fourier transformation with
$n_{\mathrm{fft}}\in\{10, 50, 100, 200, 1000\}$ components were used for feature extraction.

    \subsubsection{General experiments description}\label{sec:ged}

        Two experiments were carried out to evaluate the proposed neural network.
        
        \begin{enumerate}
            \item Regression, where the task involved modeling the hidden part of the ECG signal using
the other part. The model was trained on the part of the DS0 dataset and then tested on a
separately constructed dataset (see Section~\ref{sec:qrs}). The model training was repeated $j=24$ times,
each time with the resampled training set.
            \item Classification, where the task involved classifying cardiac arrhythmia. The model was
trained and validated using the DS1 dataset and then tested on the DS2 dataset.
Additionally, the regression-classification scenario was tested, in which the model
pretrained for the regression task was finetuned on the classification task. For both
cases, the training was repeated $j=24$ times, with resampled training,
validation and test sets.
        \end{enumerate}
        
    \subsubsection{Hyperparameters optimization experiment description}

        Model hyperparameters were obtained using Ray Tune library~\cite{Liaw2018}, and
a detailed description of the experiment’s single iteration is given below.
        \begin{enumerate}
            \item The DS0 dataset was divided into training (90\%) and validation (10\%) subsets, which
were saved and used for each iteration of the Ray Tune hyperparameter optimization.
We also chose one of 24 DS1 partitions introduced in Section~\ref{sec:ged} above, which was
used for each iteration of the Ray Tune hyperparameter optimization.
            \item The proposed CNN model was trained to perform the regression task using the DS0
training subset. After training, the DS0 validation subset was used to choose the best
network based on validation loss.
            \item The chosen network was trained for the classification task using the DS1 training
subset. After training, the best model was chosen using the DS1 validation set based on
validation loss. The value of the validation loss of the validation set was used as the
score of the model.
        \end{enumerate}
        
    \subsubsection{Classification experiment description}\label{sec:ced}
        
        Detailed steps of the classification experiment’s single iteration are given below.
        \begin{enumerate}
            \item One each of the 24 DS1 and DS2 partitions were chosen, where DS1 was further
divided into training and validation, and DS2 was used for testing.
            \item The model was trained using the training part of the DS1 dataset and the best model
was chosen using the validation subset based on validation loss.
            \item The chosen model was evaluated using the DS2 dataset.
        \end{enumerate}

    \subsubsection{Regression experiment description}\label{sec:red}
        
        Detailed steps of the regression experiment’s single iteration are given below.
        \begin{enumerate}
            \item We sampled 90\% of ECG segments from the DS0 dataset.
            \item The model was trained using the sampled ECG segments.
            \item After the predetermined number of iterations, the model was evaluated using
the prepared test dataset (see Section~\ref{section:regression}).
        \end{enumerate}

    \subsubsection{Regression-to-classification experiment description}\label{sec:rtc}
    
        Detailed steps of the regression-to-classification experiment’s single iteration are given
below.
        \begin{enumerate}
        \item We randomly divided the DS0 dataset into training (90\%) and validation (10\%)
subsets. We also chose the same DS1 and DS2 partitions used in the classification
experiment in Section~\ref{sec:ced} above.
            \item We trained the regression model using the DS0 training dataset and chose the best
model using the DS0 validation dataset based on validation loss.
            \item The best regression model was trained on the classification task using the DS1 training
subset and the best classification model was chosen using the DS1 validation subset
based on validation loss.
            \item Model performance was evaluated using the DS2 dataset.
        \end{enumerate}

\begin{figure}[!ht]
            \centering
            \includegraphics[width=0.7\columnwidth]{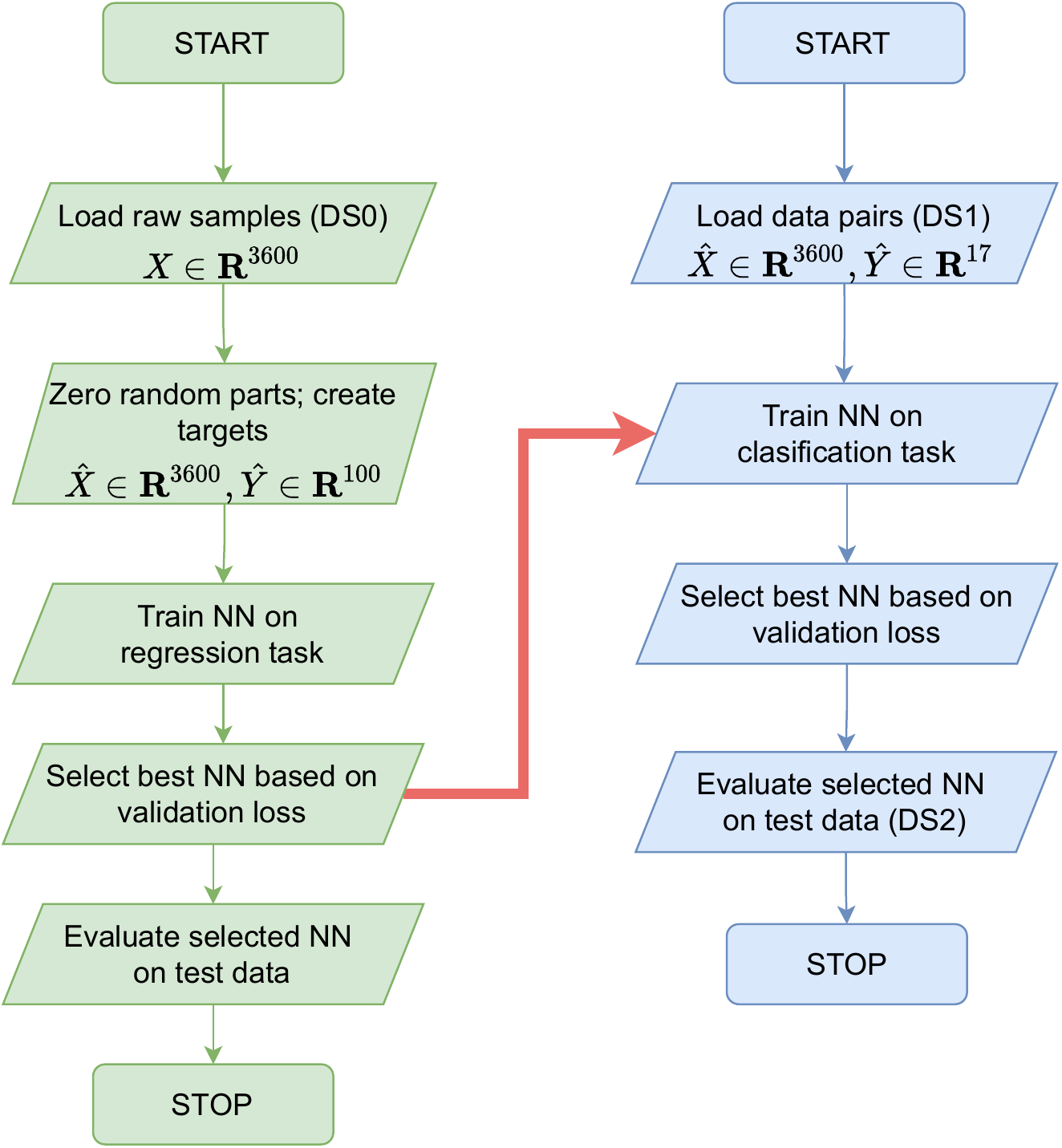}
            \caption{Flowcharts present training pipelines for regression (green) and classification (blue) experiments. Red arrow describes the combined pipeline used in the regression-to-classification experiment. In the case of regression, selecting best model based on validation loss was done only in the case of regression-to-classification experiment (see Sections~\ref{sec:red} and~\ref{sec:rtc} for details).}
            \label{figure:flowcharts}
        \end{figure}

Figure~\ref{figure:flowcharts} illustrates the workflow for the regression and classification experiments that
have been described above.

    \section{Results and Discussion}\label{section:experiments}
        The results of the proposed neural network on regression and classification tasks are
presented and discussed.

    \subsection{Regression results: ECG QRS prediction}\label{section:regression}\label{sec:qrs}
    
\begin{figure}[!ht]
	\centering
	\begin{subfigure}[b]{0.44\linewidth}
		\includegraphics[width=1\linewidth]{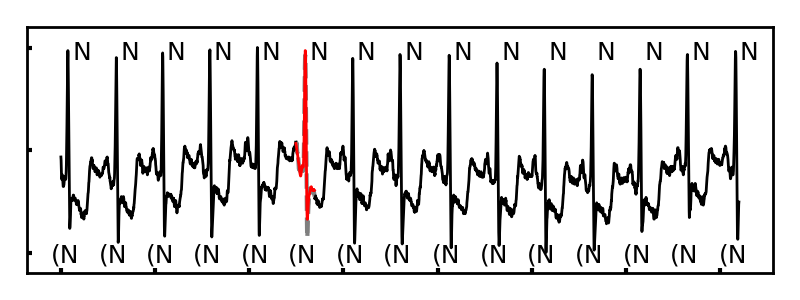}
		\caption{Example full window, `N' group}\label{fig:qrs:intro:1}
	\end{subfigure}
	\begin{subfigure}[b]{0.44\linewidth}
		\includegraphics[width=1\linewidth]{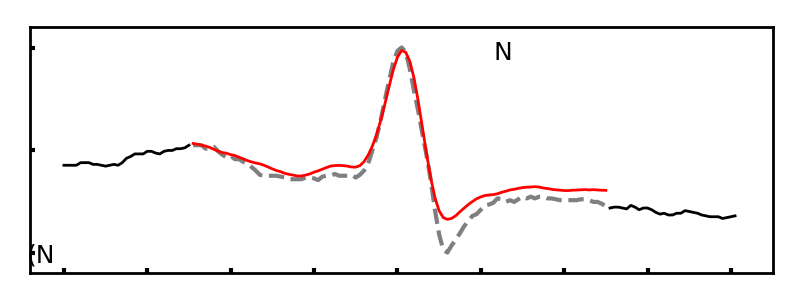}
		\caption{Zoom of (a) at zeroed (predicted) fragment}\label{fig:qrs:intro:1z}
	\end{subfigure}
	\begin{subfigure}[b]{0.44\linewidth}
		\includegraphics[width=1\linewidth]{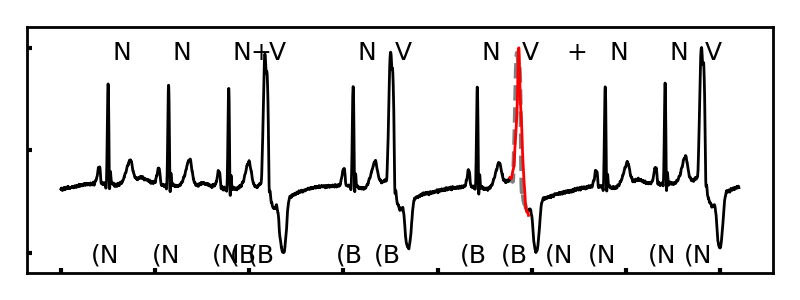}
		\caption{Example full window, `M' group}\label{fig:qrs:intro:2}
	\end{subfigure}
	\begin{subfigure}[b]{0.44\linewidth}
		\includegraphics[width=1\linewidth]{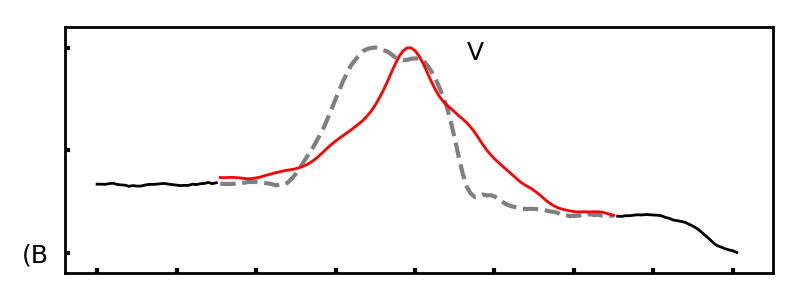}
		\caption{Zoom of (c) at zeroed (predicted) fragment}\label{fig:qrs:intro:2z}
	\end{subfigure}
	\caption{Example QRS prediction in the experiment (see Section~\ref{sec:qrs}). An example input window (a) is an ECG fragment of $n_i=3600$ samples; part of it (gray dashed line) has been zeroed and was predicted by the network (red line). Details of the prediction can be seen in the zoomed inset (b). As the (a)/(b) window contains only one class of beat and rhythm, denoted by the normal beat (`N') and normal sinus rhythm (`(N') labels, it is of group `N'. Another example input (c) is a fragment with more diversified signal; there are additionally premature ventricular contraction type beats (`V') and ventricular bigeminy rhythm ('(B') present. Since there's more than one type of pattern, it belongs to the `M' group. Note that the zeroed QRS is at different places with the input windows.}
	\label{fig:qrs:intro}
    \end{figure}

An ECG signal from the study database could be viewed as a sequence of single beats or QRS complexes, each of which had been manually annotated by experts~\cite{Moody_2001}. To answer the question of how well the individual QRS complexes could be
reproduced by the pretrained network, we designed the experiment where, repeatedly, a
portion of the ECG segment of fixed length (100 datapoints) containing at least one QRS
complex was removed and its prediction, computed by the network, was evaluated. For this
experiment, we used the original database annotations, which contained both the
classification and location of individual QRS beats~\cite{Moody_2001}. The QRS
annotations were typically centered around the peak of the R wave on the used modified
ECG Lead II signal. Selecting one QRS complex from a randomly selected ECG segment (with
fixed length of 3,600 datapoints) in the database, we zeroed a $n_z =100$ point window
centered around the R peak position. The zeroing window was shifted $\pm10\%$ to the left and
right to vary the neurons predicting different parts of the signal. Without the introduced
shift, the experiment would effectively test only a small fraction of the output neurons. On
the other hand, too much shift would risk cutting off informative parts of the QRS. The
value of $\pm 10\%$ was a reasonable compromise that balanced the computational load within the
network and preservation of the input pattern. The 100-sample zeroed window was thus
embedded at a random position within the $n_i =3600$ sample points network input window.
Each 10-second ECG segment contained 3,600 sampled datapoints, which typically
encompassed 10 to 15 QRS complexes or beats. A beat could be annotated as a normal
sinus beat “N”, a paced beat “/”, or non-sinus non-paced beat. The use of a left parenthesis
marked the underlying rhythm, e.g., “(N” and “(P” denoted normal sinus rhythm and paced
rhythm, respectively. In addition, the 10-segment ECG segments could be divided into four
groups depending on database annotations within the network input window: “N”
contained only normal sinus beats and normal sinus rhythm (i.e., “N” beat and “(N” rhythm
annotations in the database); “P” contained pacemaker signals (i.e., “/” beat and “(P”
rhythm annotations); “X” contained only a single type of beat and rhythm that were non-
normal and non-paced (i.e., beat was not “N” “/” and/or rhythm was not “(N” “(P”’); and
“M” encompassed the remaining situations with two or more beat classes and/or two or
more rhythm classes within an input window. The ground truth annotations of ECG beat
and rhythm were used to randomly select $n_n =n_x =n_m =10000$ input vectors from each of the
“N”, “X” and “M” groups as well as $n_p =2281$ vectors from available ECG segments in the “(P”
group. Figure~\ref{fig:qrs:intro} shows two examples of input windows.

For this experiment, we used $n=24$ models to train for regression on the 90\% randomly
selected subset of the DS0 dataset. After training, each member of the selected “N”, “X, “M”
and “P” groups was placed as input to one of the models, selected in a round-robin scheme.
As the test set had been selected independently of the training/validation splits, there
might exist overlap between the two. Each prediction step of the experiment followed the
regression stage, i.e., the network output a $n_o =100$ length vector that was compared with
the reference signal (the original zeroed portion). Normalized root mean square error
(NRMSE) distance between the original $\vect{y} = \begin{bmatrix}y_{1}, ..., y_{n_o}\end{bmatrix}$ and the predicted $\hat{\vect{y}} = \begin{bmatrix}\hat{y}_{1}, ..., \hat{y}_{n_o}\end{bmatrix}$ signals was used as an error measure, which is given by:
    \begin{equation}
        e_{\text{nrmse}} = \frac{\sqrt{\frac{1}{n_o} \sum_{i=1}^{n_o} (y_i - \hat{y}_{i})^2}}{\max(\vect{y}) - \min(\vect{y})}
    \end{equation}
NRMSE allowed us to represent the error of reproduction relative to the signal range.
Additionally, we measure the NRMSE of both sequences scaled to $\langle 0, 1\rangle$ (to remove
variations in amplitude), shifted by $\pm 10\%$ (best NRMSE value, to remove the effect of slight
time shift) and scaled+shifted. A total of $n_m =24$ independently-trained models were
deployed in this experiment.

\begin{figure}
    	\centering
    	\begin{subfigure}[b]{0.3\linewidth}
    		\includegraphics[width=1\linewidth]{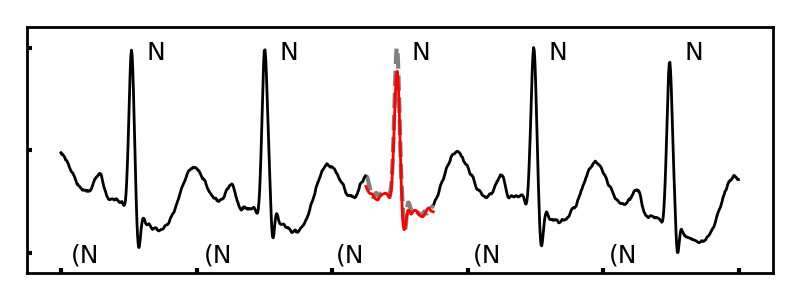}
    		\caption{N, good, $e_{\text{nrmse}}\approx0.03$}\label{fig:qrs:n:b}
    	\end{subfigure}
    	\begin{subfigure}[b]{0.3\linewidth}
    		\includegraphics[width=1\linewidth]{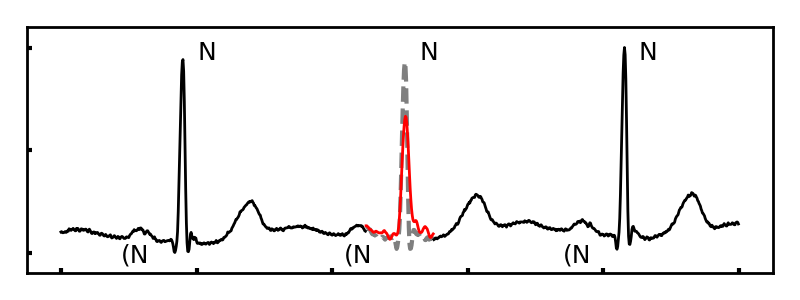}
    		\caption{N, median, $e_{\text{nrmse}}\approx0.1$}\label{fig:qrs:n:m}
    	\end{subfigure}
    	\begin{subfigure}[b]{0.3\linewidth}
    		\includegraphics[width=1\linewidth]{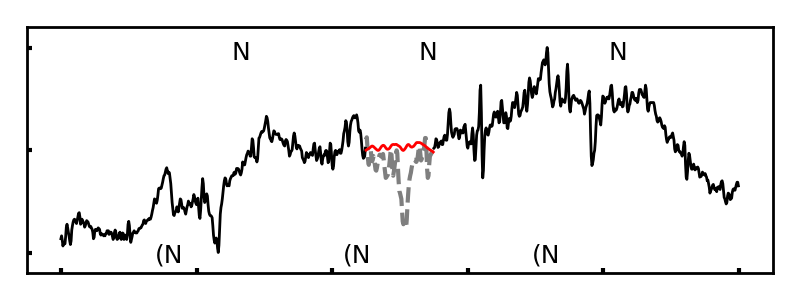}
    		\caption{N, bad, $e_{\text{nrmse}}\approx1.51$}\label{fig:qrs:n:w}
    	\end{subfigure}
    
    	\begin{subfigure}[b]{0.3\linewidth}
			\includegraphics[width=1\linewidth]{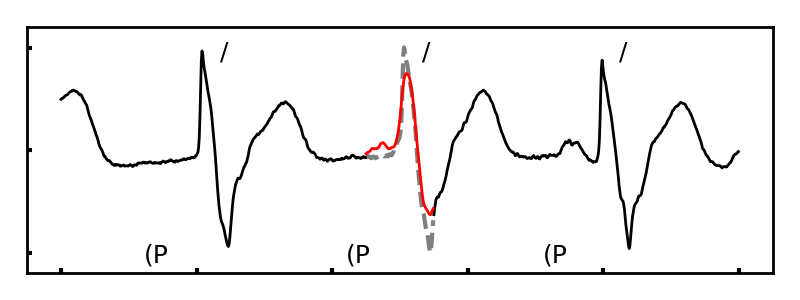}
			\caption{P, good, $e_{\text{nrmse}}\approx0.04$}\label{fig:qrs:p:b}
		\end{subfigure}
		\begin{subfigure}[b]{0.3\linewidth}
			\includegraphics[width=1\linewidth]{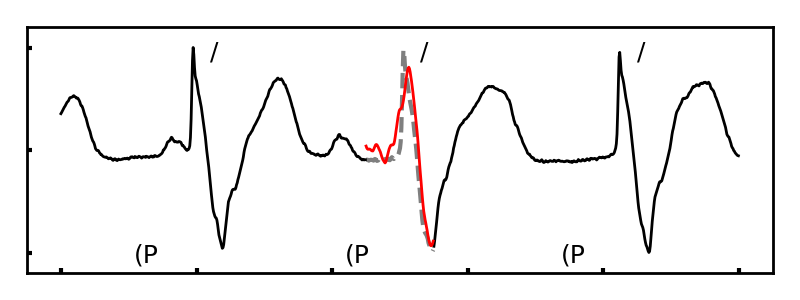}
			\caption{P, median, $e_{\text{nrmse}}\approx0.08$}\label{fig:qrs:p:m}
		\end{subfigure}
		\begin{subfigure}[b]{0.3\linewidth}
			\includegraphics[width=1\linewidth]{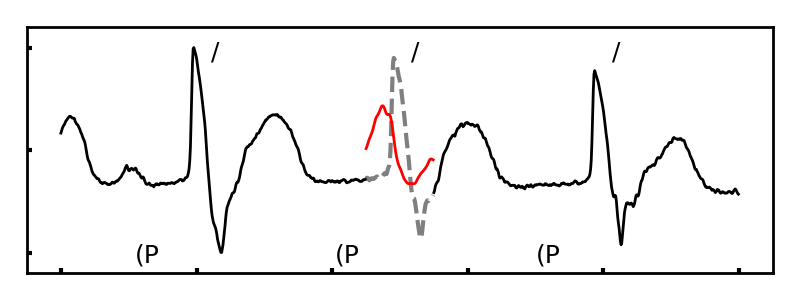}
			\caption{P, bad, $e_{\text{nrmse}}\approx0.33$}\label{fig:qrs:p:w}
		\end{subfigure}
    
    	\begin{subfigure}[b]{0.3\linewidth}
    		\includegraphics[width=1\linewidth]{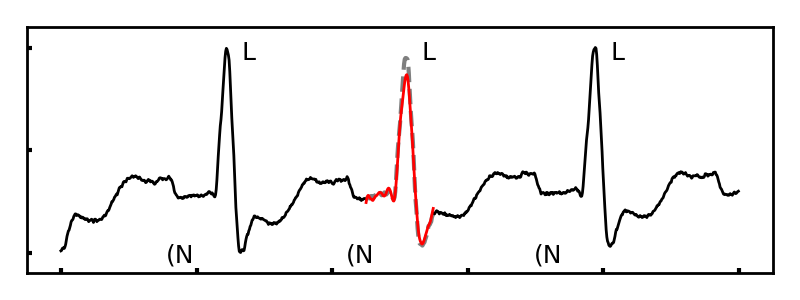}
    		\caption{X, good, $e_{\text{nrmse}}\approx0.03$}\label{fig:qrs:x:b}
    	\end{subfigure}
    	\begin{subfigure}[b]{0.3\linewidth}
    		\includegraphics[width=1\linewidth]{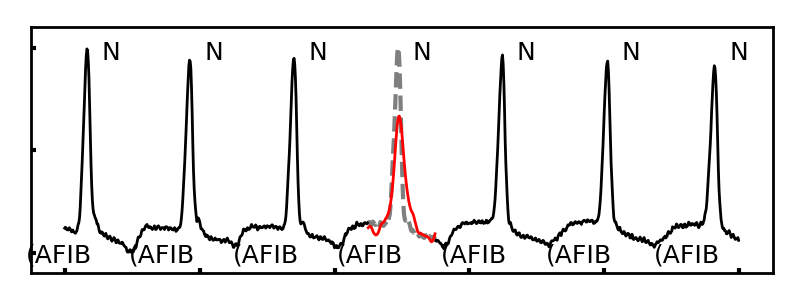}
    		\caption{X, median, $e_{\text{nrmse}}\approx0.12$}\label{fig:qrs:x:m}
    	\end{subfigure}
    	\begin{subfigure}[b]{0.3\linewidth}
    		\includegraphics[width=1\linewidth]{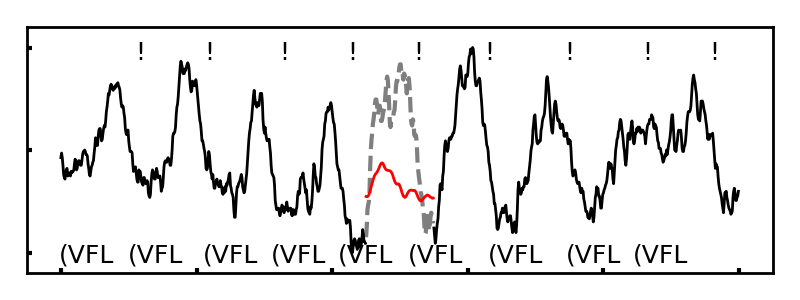}
    		\caption{X, mad, $e_{\text{nrmse}}\approx0.58$}\label{fig:qrs:x:w}
    	\end{subfigure}

    	\begin{subfigure}[b]{0.3\linewidth}
			\includegraphics[width=1\linewidth]{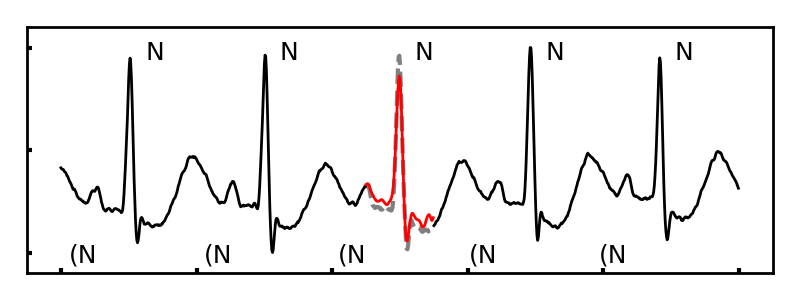}
			\caption{M, good, $e_{\text{nrmse}}\approx0.03$}\label{fig:qrs:m:b}
		\end{subfigure}
		\begin{subfigure}[b]{0.3\linewidth}
			\includegraphics[width=1\linewidth]{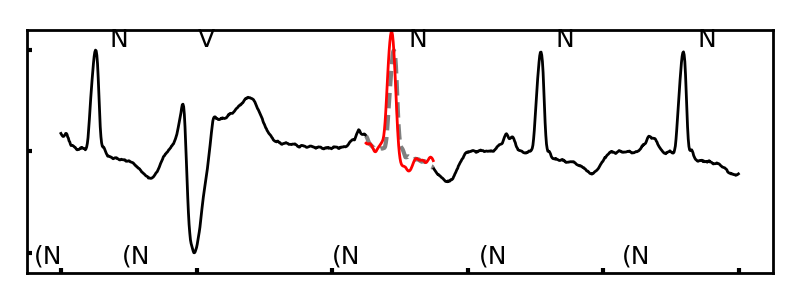}
			\caption{M, median, $e_{\text{nrmse}}\approx0.14$}\label{fig:qrs:m:m}
		\end{subfigure}
		\begin{subfigure}[b]{0.3\linewidth}
			\includegraphics[width=1\linewidth]{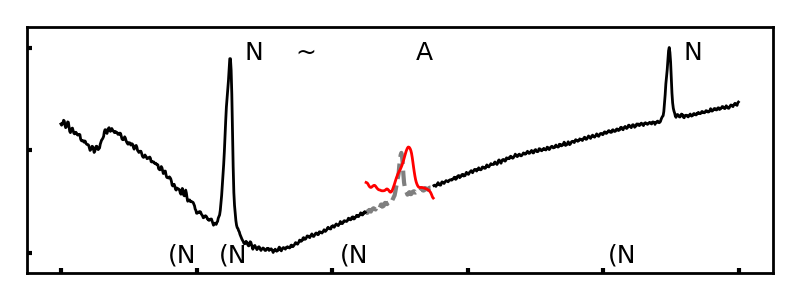}
			\caption{M, bad, $e_{\text{nrmse}}\approx2.0$}\label{fig:qrs:m:w}
		\end{subfigure}

    	\begin{subfigure}[b]{0.3\linewidth}
			\includegraphics[width=1\linewidth]{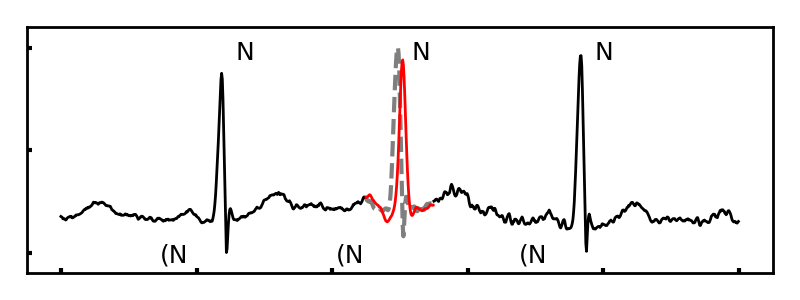}
			\caption{N, sample, $e_{\text{nrmse}}\approx0.3$}\label{fig:qrs:e:t}
		\end{subfigure}
		\begin{subfigure}[b]{0.3\linewidth}
			\includegraphics[width=1\linewidth]{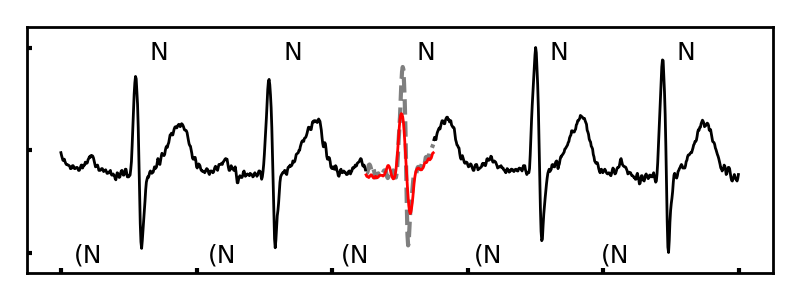}
			\caption{M, sample, $e_{\text{nrmse}}\approx0.08$}\label{fig:qrs:e:a0}
		\end{subfigure}
		\begin{subfigure}[b]{0.3\linewidth}
			\includegraphics[width=1\linewidth]{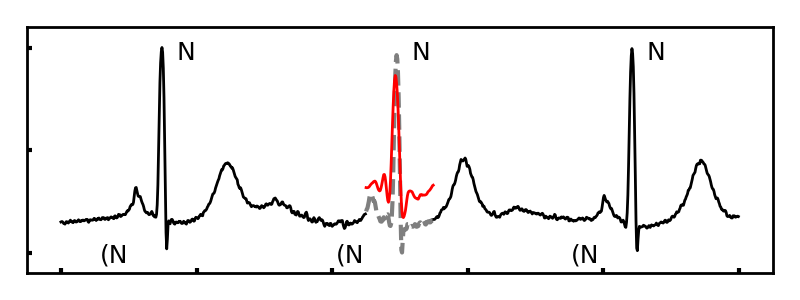}
			\caption{M, sample, $e_{\text{nrmse}}\approx0.07$}\label{fig:qrs:e:a1}
		\end{subfigure}    

    	\begin{subfigure}[b]{0.3\linewidth}
			\includegraphics[width=1\linewidth]{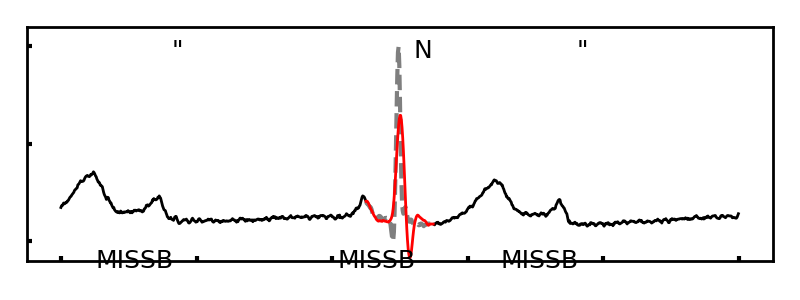}
			\caption{M, sample, $e_{\text{nrmse}}\approx0.11$}\label{fig:qrs:e:0}
		\end{subfigure}
		\begin{subfigure}[b]{0.3\linewidth}
			\includegraphics[width=1\linewidth]{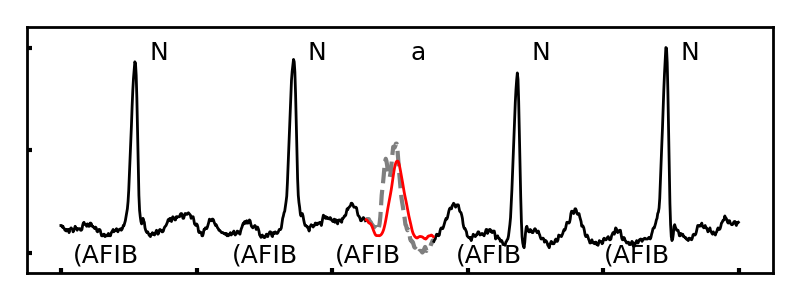}
			\caption{M, sample, $e_{\text{nrmse}}\approx0.14$}\label{fig:qrs:e:1}
		\end{subfigure}
		\begin{subfigure}[b]{0.3\linewidth}
			\includegraphics[width=1\linewidth]{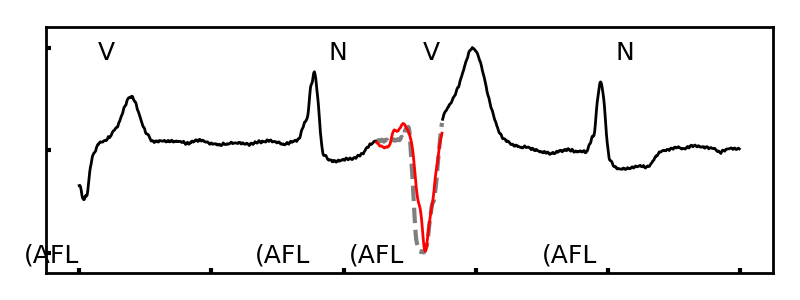}
			\caption{M, sample, $e_{\text{nrmse}}\approx0.12$}\label{fig:qrs:e:2}
		\end{subfigure}
    
    	\caption{Examples from QRS prediction experiment (see Section~\ref{sec:qrs}). Each plot presents a single prediction result, centered on the predicted signal, with the prediction marked in red, while true signal is depicted in grey dashed line. The symbols within the plots correspond to the beat and rhythm annotations in the original database. Figures~\ref{fig:qrs} (a-l) present good, median, and bad from the four groups. The following figures present examples of problematic predictions: time shift (m), amplitude scale (n) and amplitude-shift (o). The remaining figures present selected cases: prediction among missed beats (p) and irregular beats/rhythm (q, r).} 
    	\label{fig:qrs}. 
    \end{figure}

\begin{figure}
    	\centering
    	\begin{subfigure}[b]{0.3\linewidth}
    		\includegraphics[width=1\linewidth]{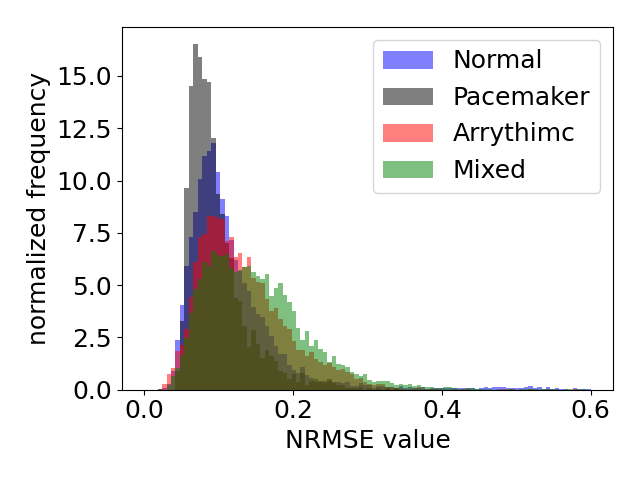}
    		\caption{Errors across four groups}\label{fig:qrs:h}
    	\end{subfigure}
    	\begin{subfigure}[b]{0.3\linewidth}
    		\includegraphics[width=1\linewidth]{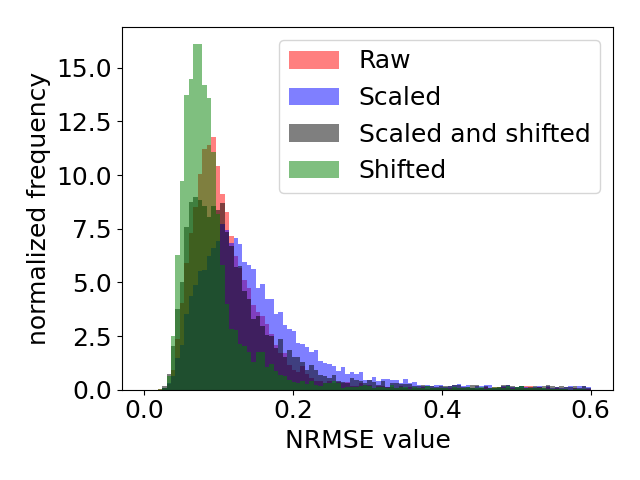}
    		\caption{The `N' group errors}\label{fig:qrs:h:n}
    	\end{subfigure}
    	\begin{subfigure}[b]{0.3\linewidth}
    		\includegraphics[width=1\linewidth]{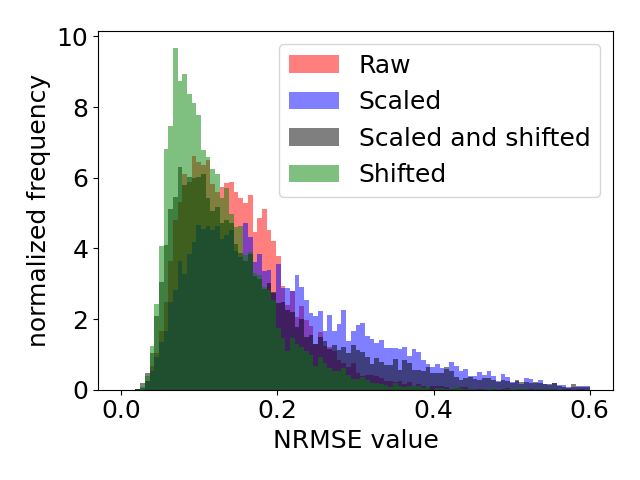}
    		\caption{The `M' group errors}\label{fig:qrs:h:m}
    	\end{subfigure}
    	\caption{Histograms of NRMSE errors from QRS prediction experiment (see Section~\ref{sec:qrs}).} 
    	\label{fig:qrs:hist}
    \end{figure}

The results are presented as plots of selected cases (Figure~\ref{fig:qrs}) and histograms of errors
(Figure~\ref{fig:qrs:hist}). The majority of cases were within $e_{\text{nrmse}} < 0.2$, which corresponded to a subjective
evaluation of moderate-to-good QRS reproduction; those QRS complexes are reproduced at
the expected positions and in shapes similar to the targets. It was observed that the network
was mainly focused on reproducing the components of the QRS wave, with less precision in
the P or T waves and other features. This may be explained by the difference in amplitudes
or numerical values, of the waves, e.g., R wave error generated much more feedback in
training than P wave error. A small number of reproductions were very accurate ($e_{\text{nrmse}} < 0.05$, Figure~\ref{fig:qrs:n:b},\ref{fig:qrs:p:b},\ref{fig:qrs:x:b},\ref{fig:qrs:m:b}). Many cases were classified as moderately
well-reproduced ($0.05 < e_{\text{nrmse}} < 0.15$, Figure~\ref{fig:qrs:n:m},\ref{fig:qrs:p:m},\ref{fig:qrs:x:m},\ref{fig:qrs:m:m}), which appeared to suffer from
three types of distortion: rescaling (the pre-training reproduced signals with different
max/min ranges and baseline shifts, see Figure~\ref{fig:qrs:e:a0},\ref{fig:qrs:e:a1}); time shift (signal characteristic
patterns were predicted earlier or later than they appeared in the signal, see Figure~\ref{fig:qrs:e:t});
and smoothing (some signal characteristics, especially high frequency details, were
smoothed out). The remaining cases ($e_{\text{nrmse}} > 0.15$,   Figure~\ref{fig:qrs:n:w},\ref{fig:qrs:p:w},\ref{fig:qrs:x:w},\ref{fig:qrs:m:w}) either did not output
QRS-like signals or estimated shapes at odds with the true data. Many of
these errors were attributable to baseline wander. Notably, the network was able to
estimate the correct shapes in many of the anticipated difficult “M” group cases, which
contained multiple different beats in sequence (see Figure~\ref{fig:qrs:e:0},\ref{fig:qrs:e:1},\ref{fig:qrs:e:2}).

Overall, the network’s ability to predict QRS could be evaluated as satisfactory for learning
the main features of the ECG signal. There was an expected variation of behavior across
groups that was apparent both in the individual cases (Figure~\ref{fig:qrs}) as well as the error
histograms (Figure~\ref{fig:qrs:h}). The least errors were seen in the “P” group, possibly because the
distinctive appearance of pacemaker signals induced less variation. The “N” group, which
had a uniform rhythm that was also well-represented in the training data, had the next best
results in terms of the error distribution. However, some cases remained problematic. The “X”
group exhibited large pattern variations and were less represented in terms of sample
number than “N”. Unsurprisingly, the mixed “M” group had the highest error due to the
added complexity of variations from different beats. A comparison of raw, scaled, shifted
and scaled+shifted NRMSE shows that shifting the zeroing window yielded the best
performance with least error (Figure~\ref{fig:qrs:h:n},\ref{fig:qrs:h:m}). In contrast, rescaling the signals prior to error
computation did not improve the performance. The amplitude of the ECG signal is not
directly related to the heart rhythm, and usually reflects other cardiac disorders, such as
atrial or ventricular hypertrophy. Some of the other observations could not be traced to the
effect of specific feature of the algorithm, e.g., it was not established whether smoothing
was due to the regularization effect of removing random noise or the lack of capacity or
inefficiency of the training procedure. Given the black box nature of artificial neural
networks, demonstrating explainability of network behavior is challenging but not
impossible~\cite{SELVARAJU_2017}.

    \subsection{Classification results}

\begin{figure}[!ht]
    	\centering
    	\begin{subfigure}[b]{0.44\linewidth}
    		\includegraphics[width=1\linewidth]{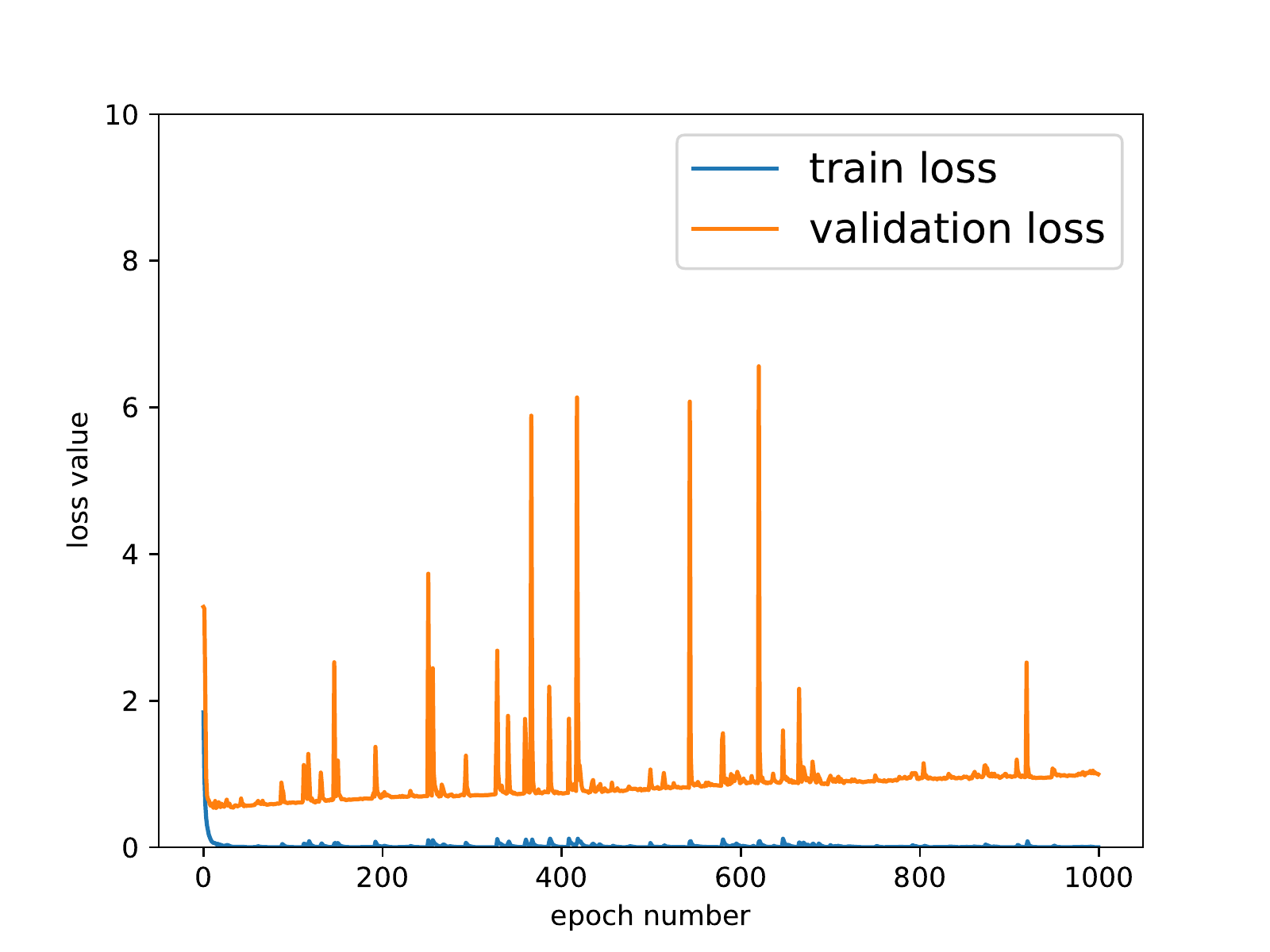}
    		\caption{Train \& validation loss curves for CNet}\label{fig:train-val-losses:c}
    	\end{subfigure}
    	\begin{subfigure}[b]{0.44\linewidth}
    		\includegraphics[width=1\linewidth]{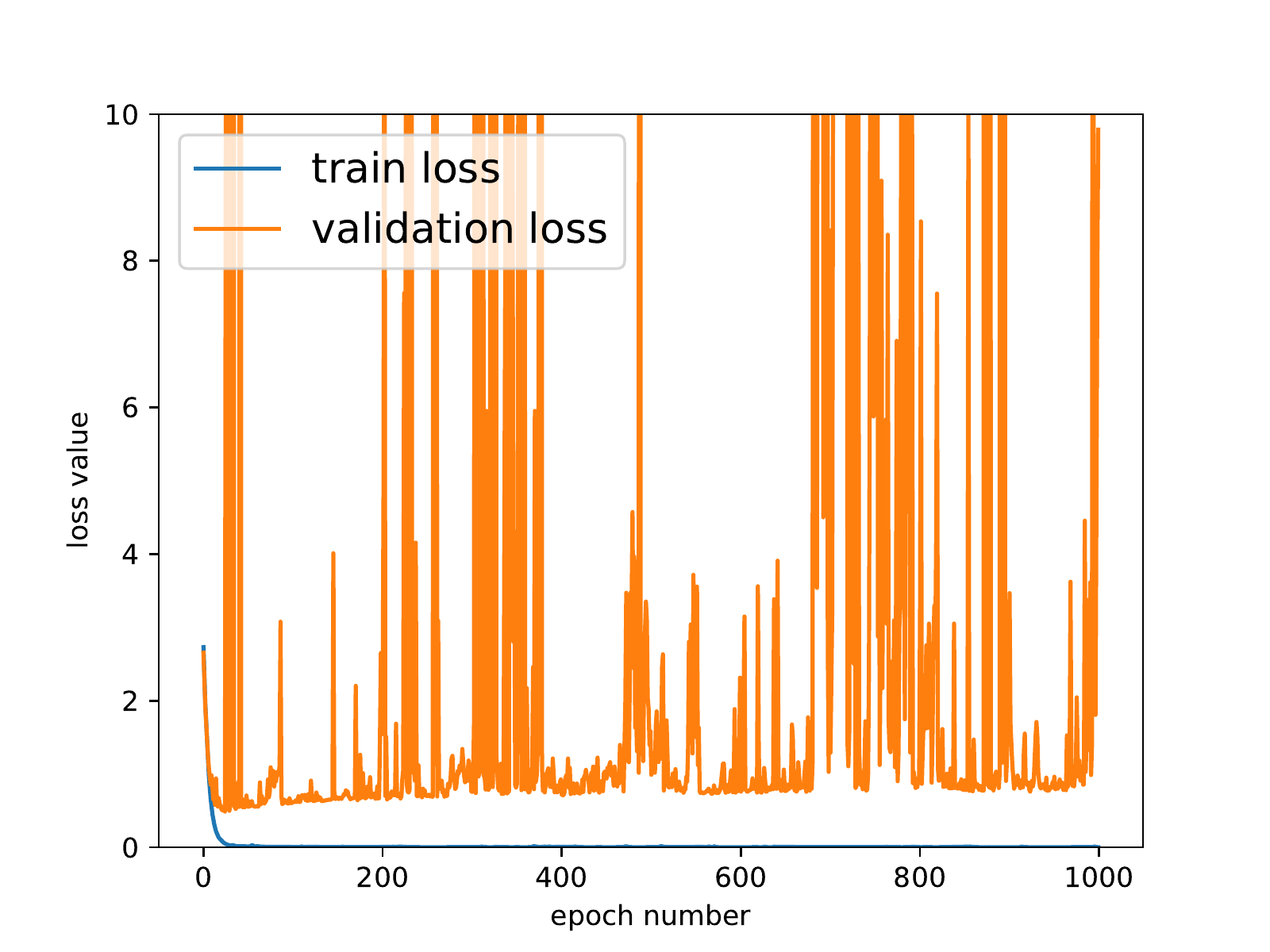}
    		\caption{Train \& validation loss curves for RCNet}\label{fig:train-val-losses:rc}
    	\end{subfigure}
    	\caption{Train \& validation loss curves for CNet (a) and RCNet (b) models (average of all 24 runs).}
    	\label{fig:train-val-losses}
        \end{figure}

The classification performance of the proposed model as well as the two reference methods
(1D-CNN and SVM, see Section~\ref{sec:ref} above) are summarized in the Table~\ref{table:classification}. CNet
denotes the proposed neural network trained for the classification task, and RCNet, the
neural network first pre-trained using the regression task and then finetuned on the
classification task. The results for our model were obtained for iterations of training, each
time using different train-validation-test splits. These same splits were used in the case of
SVM and 1D-CNN reference methods. The performance metrics of overall accuracy (ov
acc), balanced accuracy (bal acc), defined as the average of recall obtained on each class, and Cohen’s Kappa statistic (kappa), are presented as
mean score and standard deviation. Balanced accuracy allows for better interpretation of the results given imbalanced dataset, while Cohen’s Kappa statistic expresses agreement between two labelings corrected for agreement by chance. Figure~\ref{fig:train-val-losses} shows averaged training and validation loss
curves for the CNet and RCNet models, while Tables~\ref{table:conf-matrix-c} and~\ref{table:conf-matrix-rc} show the summed confusion matrices for these models.

        \begin{ctable}[
            caption={Effectiveness of the proposed neural network on the classification task.},
            label={table:classification},
            pos=h
            ]{cccc}{}
            {
            \toprule
            & ov acc (\%) & bal acc (\%) & kappa \\
            \midrule
            CNet & $87.33\pm2.5$ & $80.54\pm5.0$ & $0.85\pm0.03$  \\
            \midrule
            RCNet & $87.78\pm2.9$ & $75.87\pm6.0$ & $0.86\pm0.03$ \\
            \midrule
            1D-CNN & $89.33\pm2.2$ & $78.72\pm5.6$ & $0.88\pm0.03$ \\
            \midrule
            SVM & $78.36\pm2.8$ & $62.78\pm5.8$ & $0.75\pm0.03$ \\
            \bottomrule
            }
        \end{ctable}

        \begin{ctable}[
            caption={Sum of confusion matrices for the CNet. The classes are in the same order as in Table~\ref{tab:dataset}.},
            label={table:conf-matrix-c},
            pos=H
            ]{cc|ccccccccccccccccc}{}
            {
            &\multicolumn{1}{c}{} & \multicolumn{16}{c}{\text{Predicted classes}} \\
            & & 1 & 2 & 3 & 4 & 5 & 6 & 7 & 8 & 9 & 10 & 11 & 12 & 13 & 14 & 15 & 16 & 17 \\
            \cline{2-19}
            &1 & 774 & 49 & 0 & 0 & 1 & 0 & 31 & 1 & 0 & 0 & 0 & 0 & 6 & 1 & 1 & 0 & 0 \\
            &2 & 96 & 160 & 0 & 10 & 2 & 0 & 13 & 1 & 0 & 0 & 0 & 1 & 4 & 0 & 1 & 0 & 0 \\
            &3 & 1 & 1 & 77 & 13 & 3 & 0 & 1 & 0 & 0 & 0 & 0 & 0 & 0 & 0 & 0 & 0 & 0 \\
            &4 & 8 & 0 & 12 & 400 & 0 & 0 & 6 & 0 & 1 & 0 & 0 & 0 & 0 & 5 & 0 & 0 & 0 \\
            &5 & 1 & 15 & 5 & 10 & 11 & 0 & 0 & 1 & 0 & 0 & 0 & 0 & 0 & 5 & 0 & 0 & 0 \\
            &6 & 0 & 0 & 0 & 0 & 0 & 48 & 0 & 0 & 0 & 0 & 0 & 0 & 0 & 0 & 0 & 0 & 0 \\
            &7 & 37 & 3 & 0 & 0 & 0 & 0 & 322 & 12 & 7 & 0 & 0 & 0 & 2 & 1 & 0 & 0 & 0 \\
            &8 & 0 & 0 & 0 & 0 & 0 & 0 & 28 & 187 & 1 & 0 & 0 & 0 & 0 & 0 & 0 & 0 & 0 \\
            \smash{\rotatebox[origin=c]{90}{\text{Actual classes}}} & 9 & 0 & 0 & 0 & 3 & 0 & 0 & 4 & 1 & 16 & 0 & 0 & 0 & 0 & 0 & 0 & 0 & 0 \\
            &10 & 0 & 0 & 0 & 0 & 0 & 0 & 4 & 5 & 0 & 33 & 0 & 2 & 0 & 3 & 0 & 0 & 1 \\
            &11 & 0 & 0 & 0 & 0 & 0 & 0 & 1 & 1 & 0 & 0 & 22 & 0 & 0 & 0 & 0 & 0 & 0 \\
            &12 & 0 & 0 & 0 & 1 & 0 & 0 & 1 & 1 & 0 & 1 & 2 & 65 & 0 & 1 & 0 & 0 & 0 \\
            &13 & 6 & 0 & 0 & 1 & 1 & 0 & 5 & 0 & 0 & 0 & 0 & 0 & 11 & 0 & 0 & 0 & 0 \\
            &14 & 1 & 0 & 0 & 1 & 0 & 0 & 2 & 0 & 0 & 0 & 0 & 0 & 0 & 451 & 0 & 0 & 1 \\
            &15 & 0 & 2 & 0 & 0 & 0 & 0 & 3 & 2 & 0 & 0 & 0 & 1 & 1 & 0 & 207 & 0 & 0 \\
            &16 & 0 & 0 & 0 & 0 & 0 & 0 & 0 & 0 & 0 & 0 & 0 & 0 & 0 & 0 & 0 & 24 & 0 \\
            &17 & 0 & 0 & 0 & 0 & 0 & 0 & 0 & 0 & 0 & 0 & 0 & 0 & 0 & 0 & 0 & 0 & 336 \\
            \cline{2-19}
            }
        \end{ctable}

        \begin{ctable}[
            caption={Sum of confusion matrices for the RCNet. The classes are in the same order as in Table~\ref{tab:dataset}.},
            label={table:conf-matrix-rc},
            pos=H
            ]{cc|ccccccccccccccccc}{}
            {
            &\multicolumn{1}{c}{} & \multicolumn{16}{c}{\text{Predicted classes}} \\
            & & 1 & 2 & 3 & 4 & 5 & 6 & 7 & 8 & 9 & 10 & 11 & 12 & 13 & 14 & 15 & 16 & 17 \\
            \cline{2-19}
            &1 & 824 & 15 & 1 & 5 & 0 & 0 & 18 & 0 & 0 & 0 & 0 & 0 & 0 & 1 & 0 & 0 & 0 \\
            &2 & 79 & 182 & 0 & 7 & 3 & 0 & 14 & 1 & 0 & 0 & 0 & 0 & 2 & 0 & 0 & 0 & 0 \\
            &3 & 2 & 3 & 62 & 14 & 7 & 0 & 5 & 0 & 0 & 0 & 0 & 0 & 1 & 2 & 0 & 0 & 0 \\
            &4 & 1 & 1 & 8 & 412 & 1 & 0 & 5 & 0 & 1 & 0 & 0 & 0 & 0 & 3 & 0 & 0 & 0 \\
            &5 & 2 & 21 & 3 & 9 & 5 & 0 & 3 & 0 & 0 & 0 & 0 & 0 & 0 & 5 & 0 & 0 & 0 \\
            &6 & 0 & 0 & 0 & 0 & 0 & 47 & 0 & 1 & 0 & 0 & 0 & 0 & 0 & 0 & 0 & 0 & 0 \\
            &7 & 37 & 4 & 0 & 3 & 0 & 0 & 317 & 13 & 7 & 0 & 0 & 0 & 0 & 3 & 0 & 0 & 0 \\
            &8 & 0 & 2 & 0 & 1 & 0 & 0 & 33 & 172 & 4 & 0 & 0 & 0 & 0 & 4 & 0 & 0 & 0 \\
            \smash{\rotatebox[origin=c]{90}{\text{Actual classes}}} &9 & 0 & 0 & 0 & 0 & 0 & 0 & 4 & 0 & 17 & 0 & 0 & 0 & 0 & 3 & 0 & 0 & 0 \\
            &10 & 0 & 0 & 0 & 0 & 0 & 0 & 3 & 4 & 1 & 37 & 0 & 1 & 0 & 0 & 1 & 0 & 1 \\
            &11 & 0 & 0 & 0 & 0 & 0 & 0 & 2 & 1 & 0 & 0 & 21 & 0 & 0 & 0 & 0 & 0 & 0 \\
            &12 & 0 & 2 & 3 & 0 & 0 & 0 & 3 & 5 & 3 & 6 & 5 & 37 & 1 & 2 & 4 & 0 & 1 \\
            &13 & 4 & 0 & 0 & 3 & 1 & 0 & 11 & 0 & 1 & 0 & 0 & 0 & 4 & 0 & 0 & 0 & 0 \\
            &14 & 0 & 0 & 0 & 0 & 1 & 0 & 3 & 1 & 0 & 0 & 0 & 0 & 0 & 451 & 0 & 0 & 0 \\
            &15 & 0 & 0 & 0 & 0 & 0 & 0 & 3 & 0 & 0 & 0 & 0 & 0 & 0 & 0 & 213 & 0 & 0 \\
            &16 & 0 & 0 & 0 & 0 & 0 & 0 & 0 & 0 & 0 & 0 & 0 & 0 & 0 & 0 & 0 & 24 & 0 \\
            &17 & 0 & 0 & 1 & 0 & 0 & 0 & 0 & 0 & 0 & 0 & 0 & 0 & 0 & 0 & 0 & 0 & 335 \\
            \cline{2-19}
            }
        \end{ctable}

Our proposed model, CNet, attained good results compared with other methods. The results were comparable with 1D-CNN and better than SVM algorithm. Compared with 1D-CNN, our model attained worse results in terms of overall accuracy and Cohen’s kappa, but was more effective when considering the balanced accuracy score, which is important here because the considered dataset was very imbalanced: some test set classes contained only one sample, while the most populous class contained 36 samples (Table~\ref{tab:dataset}).

Our second approach, RCNet, attained comparable results to the original CNet model. It was similar in terms of overall accuracy and Cohen’s kappa, but was worse when considering balanced accuracy. The lack of improvement compared with CNet could be explained by multiple factors. The predictive performance of the regression, while generally good, might not have been at a high enough level. On the other hand, it is also possible that the regression training resulted in overfitting to this specific task, making the model less useful as a pretraining tool.

Overall, the results indicate the high effectiveness of the proposed method for the task of classifying heart arrhythmia, especially in the case of imbalanced datasets.

    \section{Conclusions}\label{sec:conclusions}

    In this article, we proposed a new deep neural network model for the classification of cardiac arrhythmia, as well as self-supervised prediction of the ECG beat signal. The same architecture was able to solve both of these tasks with minimal adjustments. We conducted experiments that confirm the effectiveness of the proposed approach, which combined unsupervised as well as supervised data to effectively train the neural network. We conclude that the proposed method is an efficient tool for the problem of cardiac arrhythmia classification based on ECG signal and the task of ECG signal approximation.

    In the future, more datasets, as well as more deep learning methods can be examined. Moreover, the joined training of regression and classification tasks can be inspected in more detail, e.g. by utlizing multi-task learning framework, in which both tasks can be learned at the same time, as opposed to training the model one task at a time.

    \section*{Acknowledgements}
    
    B.G. acknowledges funding from the budget funds for science in the years 2018-2022, as a scientific project ''Application of transfer learning methods in the problem of hyperspectral images classification using convolutional neural networks'' under the ''Diamond Grant'' program, no. DI2017 013847.

    \bibliographystyle{plain}
    \bibliography{references}
\end{document}